\documentstyle[11pt]{article}

\makeatletter

\newcommand{\CiteLeftDelim}{(}

\newcommand{\CiteRightDelim}{)}

\newcommand{\CiteDelimsParens}{%
\renewcommand{\CiteLeftDelim}{(}%
\renewcommand{\CiteRightDelim}{)}}

\newcommand{\CiteDelimsEmpty}{%
\renewcommand{\CiteLeftDelim}{}%
\renewcommand{\CiteRightDelim}{}}

\newcommand{\CiteListSeparator}{;}
\newcommand{\CiteListSemicolon}{\renewcommand{\CiteListSeparator}{;}}
\newcommand{\CiteListComma}{\renewcommand{\CiteListSeparator}{,}}

\def\@citex[#1]#2{\if@filesw\immediate\write\@auxout{\string\citation{#2}}\fi%
\def\@citea{}\@cite{\@for\@citeb:=#2\do%
{\@citea\def\@citea{\CiteListSeparator\penalty\@m\ }\@ifundefined%
{b@\@citeb}{{\bf ?}\@warning%
{Citation `\@citeb' on page \thepage \space undefined}}%
{\csname b@\@citeb\endcsname}}}{#1}}
\let\@internalcite\cite
\def\cite{\CiteListSemicolon\CiteDelimsParens\def\citename##1{##1
	}\@internalcite}
\def\shortcite{\CiteListComma\CiteDelimsParens\def\citename##1{}\@internalcite}
\def\newcite{\CiteListComma\CiteDelimsParens\leavevmode\def\citename##1{{##1}
	(}\@internalciteb}
\def\ecite{\CiteListSemicolon\CiteDelimsEmpty\def\citename##1{##1
	}\@internalcite}
\def\eshortcite{\CiteListComma\CiteDelimsEmpty\def\citename##1{}\@internalcite}

\def\@citexb[#1]#2{\if@filesw\immediate\write\@auxout{\string\citation{#2}}\fi
\def\@citea{}\@newcite{\@for\@citeb:=#2\do%
{\@citea\def\@citea{\CiteListSeparator\penalty\@m\ }\@ifundefined%
{b@\@citeb}{{\bf ?}\@warning%
{Citation `\@citeb' on page \thepage \space undefined}}
\hbox{\csname b@\@citeb\endcsname}}}{#1}}%
\def\@internalciteb{\@ifnextchar [{\@tempswatrue\@citexb}
	{\@tempswafalse\@citexb[]}}

\def\@newcite#1#2{{#1\if@tempswa\CiteListSeparator #2\fi)}}

\def\@biblabel#1{\def\citename##1{##1}[#1]\hfill}

\def\@cite#1#2{\CiteLeftDelim{#1\if@tempswa\CiteListSeparator
	#2\fi}\CiteRightDelim}

\def\thebibliography#1{\section*{Bibliography\@mkboth
 {Bibliography}{Bibliography}}\list
 {}{\setlength{\labelwidth}{0pt}\setlength{\leftmargin}{20pt}
 \setlength{\itemindent}{-20pt}}
 \def\newblock{\hskip .11em plus .33em minus -.07em}
 \sloppy\clubpenalty4000\widowpenalty4000
 \sfcode`\.=1000\relax}

\def\@lbibitem[#1]#2{\item[]\if@filesw
      { \def\protect##1{\string ##1\space}\immediate
        \write\@auxout{\string\bibcite{#2}{#1}}\fi\ignorespaces}}

\def\@bibitem#1{\item\if@filesw \immediate\write\@auxout
       {\string\bibcite{#1}{\the\c@enumi}}\fi\ignorespaces}

\makeatother

\newcommand{\pf}[1]{\noindent{\bf proof }  {#1} \hfill~$\Box$}
\newcommand{\implies}{\: \supset \:}
\newcommand{\And}{\: \wedge \:}
\newcommand{\arr}{-\!\!\!\!\rightarrow}   			
\newcommand{\larr}{\leftarrow\!\!\!\!-}				
\newcommand{\darr}{-\!\!\!\!\rightarrow\!\!\!\!\!\rightarrow}   
\newcommand{\invdarr}{\leftarrow\!\!\!\!\!\leftarrow\!\!\!\! -} 
\newcommand{\drpl}{\invdarr\!\!\!\!\!\!\darr}   		

\newtheorem{defi}{Definition}
\newtheorem{LEMMA}{Lemma}
\newtheorem{COROLLARY}{Corollary}
\newtheorem{thm}{Theorem}

\newcommand{\lemma}[1]{\begin{LEMMA} {\rm {#1}} \end{LEMMA}}
\newcommand{\corollary}[1]{\begin{COROLLARY} {\rm {#1}} \end{COROLLARY}}
\newcommand{\theorem}[1]{\begin{thm} {\rm {#1}} \end{thm}}
\newcommand{\kases}{    
   \vspace{-3.2mm}
   \begin{list}
 {???}
 {\setlength{\leftmargin}{4.1mm}
  \setlength{\labelwidth}{2mm}}
  \setlength{\parsep}{1mm}
  \setlength{\itemsep}{0.01mm}
  \setlength{\topsep}{0.01mm}
  \setlength{\parskip}{0mm}
  \setlength{\parindent}{0mm}}
\newcommand{\Endkases}{\end{list}\vspace{-3.2mm}}
\newcommand{\kase}[1]{\item[case {#1}:\ ]}

\newcommand{\State}[5]
{\mbox{\vbox{state #1:\\
\parbox[t]{2em}{\verb+ + B:} \parbox[t]{\ExampleWidth}{#2}\\
\parbox[t]{2em}{\verb+ + S:} \parbox[t]{\ExampleWidth}{#3}\\
\parbox[t]{2em}{\verb+ + I:} \parbox[t]{\ExampleWidth}{#4}\\
\parbox[t]{2em}{\verb+ + P:} \parbox[t]{\ExampleWidth}{#5}
}}}

\newcommand{\assigned}{\mbox{\(\: : =\: \)}}
\newcommand{\ruul}{\mbox{\(\:\triangleright\:\)}}
\newcommand{\wate}{\NN}
\newcommand{\skore}{\mbox{{$\sigma$}}}

\newcommand{\drm}{\mbox{{$d_{\mbox{rm}}$}}}
\newcommand{\NN}{\mbox{\#}}
\newcommand{\LeftChild}{\mbox{\(\lambda\)}}
\newcommand{\RightChild}{\mbox{\(\rho\)}}

\newcommand{\suchthat}{such that}
\newcommand{\ctr}{\mbox{{\sf ctr}}}
\newcommand{\AlgorithmTabStops}{
\hspace{1in} \= 5. \ \= \ \ \ \ \= \ \ \ \ \= \ \ \ \ \= \kill}

\title{The complexity of normal form rewrite sequences for
Associativity.\thanks{This paper appears as Technical Report LCL 94-6 at the
Computer Science Department of the Technion -- Israel Institute of Technology.
It is also electronically archived in the Computation and Language E-Print
Archive as cmp-lg/9406030.}}
\date{December 22, 1993}
\author{Michael Niv}

\begin{document}
\maketitle

\section*{Abstract}

The complexity of a particular term-rewrite system is considered: the rule of
associativity \((x*y)*z \;\ruul\; x*(y*z)\).  Algorithms and exact
calculations are given for the longest and shortest sequences of applications
of \ruul\ that result in normal form (NF). The shortest NF sequence for a term
$x$ is always $n-\drm(x)$, where $n$ is the number of occurrences of $*$ in
$x$ and $\drm(x)$ is the depth of the rightmost leaf of $x$.  The longest NF
sequence for any term is of length $n(n-1)/2$.

\section{Preliminaries}

\cite{Klop92} provides an overview of the theory of term rewrite systems.
There is relatively little known about the complexity of various term rewrite
systems.  Here, I consider a particular rewrite system with one binary
connective $*$ and one rewrite rule $(x*y)*z \;\ruul\; x*(y*z)$.

A rewrite system $\arr$ is {\em Strongly Normalizing} (SN) iff every sequence
of applications of $\arr$ is finite. A rewrite system is {\em Church-Rosser
(CR)} just in case \[ \forall x,y . (x \drpl y \implies \exists z . (x \darr z
\And y \darr z)) \] A rewrite system is {\em Weakly Church-Rosser (WCR)} just
in case \[ \forall x,y,w . (w \arr x \And w \arr y) \implies \exists z . (x
\darr z \And y \darr z) \]

Let $\arr$ be the relation between two terms such that $x\arr y$ just in case
$x$ contains a subterm, the {\em redex}, which matches the left hand side of
the rule \ruul, and replacing the redex by the corresponding right hand side,
the {\em contractum}, yields the new term $y$.  A term is in {\em normal form
(NF)\/} if it contains no redex.  Let $\larr$ be the converse of $\arr$.  Let
$\longleftrightarrow$ be $ \arr \cup \larr$.  Let $\darr$ be the reflexive
transitive closure of $\arr$ and similarly, $\invdarr$ the reflexive
transitive closure of $\larr$, and $\drpl$ the reflexive transitive closure of
$\longleftrightarrow$.  Note that $\drpl$ is an equivalence relation.

Given a term $x$, define \LeftChild$(x)$ (resp.\ \RightChild$(x)$) refers to
its the left (right) child of $x$.

\section{Longest rewrite sequence}

Given a term $x$, $\wate x$ and $\skore x$ are defined as follows:

\[
\begin{array}{lll}
\wate x & \!\!=\!\! & \left\{
\begin{array}{ll}
0 \mbox{\ if $x$ is a leaf node}&\\
 1 + \wate \mbox{\LeftChild($x$)} + \wate \mbox{\RightChild($x$)} &
\mbox{otherwise}
\end{array}\right. \\ 
\mbox{\rule{0pt}{.1mm}} & & \\
\skore x & \!\!=\!\! & \left\{
\begin{array}{ll}
0 \mbox{\ if $x$ is a leaf node} &\\
\skore \mbox{\LeftChild($x$)}+\skore \mbox{\RightChild($x$)}+\wate
	\mbox{\LeftChild($x$)}&
			\mbox{\hspace{-0.5em}otherwise}
\end{array}\right. 
\end{array}
\]

Note that $\wate x$ is the number of internal nodes in $x$.  By convention,
$n$ is $\wate x$.  Note also that if $x\ruul x'$ then \mbox{\(x'=
\LeftChild(\LeftChild(x))*(\RightChild(\LeftChild(x))*\RightChild(x))\)}
and \mbox{\(\skore x' = \skore x - (\wate \LeftChild(\LeftChild(x)) + 1)\)}.

\lemma{\label{SN}$\arr$ is SN.}

\pf{Every term $x$ is assigned a positive integer measure $\skore x$.
An application of $\arr$ is guaranteed to lower the measure.  This follows
from monotonic dependency of $\skore x$ upon the \skore's of each of $x$'s
subterm, and from the fact that locally, applying \ruul\ lowers \skore\ .}

\theorem{For every term $x$, there exists a NF-yielding sequence of
$\skore x$ applications of $\arr$, furthermore, this is the longest possible
NF sequence for $x$.}

\pf{The sequence of constructed by induction on $\skore x$:

Base case: $\skore x = 0$. For every subterm $y$ of $x$, $\wate
\LeftChild(y)=0$, i.e.\ \LeftChild(y) is a leaf. So $x$ is in NF.

Induction: I show that for every term $x$ \suchthat\ $\skore(x)>0$ there
exists another term $x'$ \suchthat\ $x\arr x'$ and $\skore x' = \skore x - 1$.
Let $y$ be the deepest leftmost descendant of $x$ such that $y$ is a redex.
Note that $\LeftChild(\LeftChild(y))$ is a leaf (otherwise $\LeftChild(y)$
would be a deeper leftmost descendant redex). Let $y'$ \suchthat\ $y\ruul y'$.
So $\wate y' = \wate y$, $\skore y' = \skore y - 1$ and
by the straightforward dependency of $\skore x'$ on the \skore's of each of
$x'$'s subterms, in particular $y'$, $\skore x' = \skore x - 1$.

The maximality of the length of the rewrite sequence follows from the fact
that the applications of $\arr$ decrease \skore\ by the minimum amount
possible, 1.
}

\corollary{\label{arr-upper-bound} For every term $x$, every
sequence of applications of $\arr$ is of length at most $n(n-1)/2$.}

\pf{It suffices to show that for every term $x$, $\skore x \leq n(n-1)/2.$
By induction on $n$:

Base case: $n=1$, $\skore x = 0$.

Induction: Suppose true for all terms $x'$ \suchthat\ $\wate x'<n$.
Let $m=\NN\LeftChild(x)$. So $0 \leq m \leq n-1$ and $\NN\RightChild(x)=n-m-1$.
\begin{eqnarray*}
{\skore x-n(n-1)/2} & = & \skore \LeftChild(x)+\skore \RightChild(x)+\wate
	\LeftChild(x)-n(n-1)/2\\
                 & \leq & \frac{m(m-1)}{2} + \frac{(n-m-1)(n-m-2)}{2} + m -
	\frac{n(n-1)}{2}\\
                    & = & (m+1)(m-(n-1))\\
                 & \leq & 0 \mbox{\hspace{3em} recalling that $0 \leq m
	\leq n-1$}
\end{eqnarray*}
}

\corollary{\label{tight} There exists a term $x$ that can be rewritten to
NF by a sequence of exactly $n(n-1)/2$ applications of $\arr$.}

\pf{

An $n$-left-chain is defined as follows: A $0$-left-chain is a leaf.  An
$n$-left-chain is an\\
\mbox{$(n-1) -$left-chain $*$ a leaf.}
Let $x$ be an $n$-left-chain. $\wate x = n$.
I show by induction on $n$ that $\skore x = n(n-1)/2$:

Base case: $n=1$, $\skore x = 0$.

Induction: Suppose true for an $(n-1)-$left-chain.
\begin{eqnarray*}
\skore x & = & \skore \LeftChild(x) + \wate \LeftChild(x) \\
         & = & (n-1)(n-2)/2 + n-1\\
         & = & n(n-1)/2
\end{eqnarray*}

}

\section{Shortest rewrite sequence}

I now show that a NF of a term (in fact {\em the\/} NF) can be
computed in linear time.

\lemma{\label{l:wcr} $\arr$ is WCR.}

\pf{      

Let $w$ be a term with two distinct redexes $x$ and $y$, yielding
the two distinct terms $w'$ and $w''$ respectively.  There are a few
possibilities:
(without loss of generality, suppose $x$ is not a subterm of $y$.)
\kases
\kase{1} $y$ is either not a subterm of $x$ or it is a
subterm of $\LeftChild(x)$ or a subterm of $\RightChild(x)$ or it is
$\RightChild(x)$. In each case is clear that the order
of application of $\arr$ makes no difference.
\kase{2} $y = \LeftChild(x)$.  For convenience let $x = ((a*b)*c)*d$.
Applying \ruul\ at $x$ gives $(a*b)*(c*d)$; applying \ruul\ at $y$ gives
$(a*(b*c))*d$.  The former can be rewritten to $a*(b*(c*d))$ using one
application of \ruul, and the latter is rewritten first to $a*((b*c)*d)$
which is then rewritten to $a*(b*(c*d))$.
\Endkases
}

\lemma{\label{newman} (Newman)
$\mbox{WCR} \And \mbox{SN} \implies \mbox{CR}$.}

\lemma{\label{unique}
\(\mbox{CR} \And \mbox{SN} \implies
(\forall x,y . (x \drpl y \And x,y \mbox{ are NFs }) \implies x=y).
\)}

\theorem { $\arr$ NFs are unique.}

\pf{Follows from lemmas \ref{SN}, \ref{l:wcr} \ref{newman}, and \ref{unique}.}

Therefore any deterministic computational path of applying $\arr$ will lead to
the NF.  I now give an algorithm \ctr\ for computing NFs.  It applies \ruul\
as {\sf c}lose as possible {\sf t}o the {\sf r}oot of its argument.

\vbox{
\begin{tabbing}
\AlgorithmTabStops
\>$\ctr_1(x) $ \hspace{1in} {\sl first version}\\
\>1. \> $y \assigned x$\\
\>2. \> while $y$ is not a leaf and $\LeftChild(y)$ is a leaf\\
\>3. \> \> $y \assigned \RightChild(y)$\\
\>4. \> if $y$ is not a leaf \\
\>5. \> \> then apply \ruul\ to $y$
\end{tabbing}
}

\lemma{\label{rightChain}
The depth of the rightmost leaf of $x$ is $n$ iff $x$ is a NF}
\pf{$x$ must be an $n$-right-chain --- the mirror of an $n$-left-chain.}

\lemma{\label{drmIncreases}
If $\drm(x)<n$, algorithm $\ctr_1$ increases $\drm(x)$ by 1.}
\pf{algorithm $\ctr_1$ scans down the path from the root to the
rightmost leaf, stopping at a redex $y$. By applying $\ruul$, it pushes
everything in $\RightChild(y)$ (including the rightmost leaf) one arc further
away from the root.}

So iterating $\ctr_1$ $n-\drm(x)$ times computes the NF.  This process is
inefficient, as it needlessly rescans the prefix of its argument.  The
following algorithm avoids this inefficiency.

\vbox{
\begin{tabbing}
\AlgorithmTabStops
\>\ctr($x$) \hspace{1in} {\sl final version}\\
\>1. \> $y \assigned x$\\
\>2. \> while $y$ is not a leaf\\
\>3. \>  \> while $y$ is not a leaf and $\LeftChild(y)$ is a leaf \\
\>4. \>  \>  \> $y \assigned \RightChild(y)$\\
\>5. \>  \> if $y$ is not a leaf \\
\>6. \>  \> then apply \ruul\ to $y$
\end{tabbing}}

\theorem{\label{ctrCorrect}
Given term $x$, $\ctr(x)$ computes a NF for $x$ in $n - \drm(x)$
applications of $\arr$.}
\pf{Clearly, \ctr\ gives the same result as $\ctr_1$ run $n - \drm(x)$ times,
	that is, the NF of $x$.  }

\section{Application}

Hepple and Morrill (1989)\nocite{HepMor89} proposed using normal forms for
overcoming certain difficulties with the parsing of Combinatory Categorial
Grammar, a formalism for natural language syntax. The results above have been
incorporated into an efficient parsing algorithm (Niv 1993,
1994)\nocite{NivThesis}\nocite{Niv94}.


\end{document}